\documentclass{article}

\usepackage[sglblindworkshop, final]{neurips_2026}
\workshoptitle{AgenticOS: Co-designing Systems and ML Foundations of an OS Layer for Agentic AI}

\usepackage[utf8]{inputenc}
\usepackage[T1]{fontenc}
\usepackage{amsmath}
\usepackage{amssymb}
\usepackage{booktabs}
\usepackage{enumitem}
\usepackage{flafter}
\usepackage{graphicx}
\usepackage[draft]{hyperref}
\usepackage{microtype}
\usepackage{multirow}
\usepackage{nicefrac}
\usepackage{url}
\usepackage{xcolor}
\usepackage{xspace}

\setcitestyle{numbers,square,comma}

\newcommand{\revise}{\textsc{Revise}\xspace}

\title{REVISE: Validity-Guided Recovery for Online Revisions in Agent Workflows}

\author{%
   Ruoling Qi$^{1,2}$ \quad Xuaner Wu$^{1}$ \quad Penghang Liu \quad Jian Chen \quad Yirui Liu$^{1,\dagger}$\\
  $^{1}$Institute of Artificial Intelligence, China Telecom (TeleAI) \\
  $^{2}$Shanghai Jiao Tong University  \\
  $^{\dagger}$Corresponding author \\
   Ruoling Qi: \texttt{qiruoling760@sjtu.edu.cn} \quad Yirui Liu: \texttt{yiruiliu926@gmail.com}}

\begin{document}

\maketitle

\begin{abstract}
Agent revisions expose a fundamental correctness--efficiency trade-off during concurrent execution. Discarding ongoing work preserves latest-version correctness but wastes progress that may remain valid, whereas reusing prior work preserves efficiency but risks propagating stale state into outputs and tool effects. Existing recovery strategies resolve this trade-off in an imbalanced way with coarse-grained policies: they either favor efficiency by allowing potentially stale work to continue, or favor correctness by restarting the workflow or recomputing a linear suffix from the earliest conflict, thereby discarding unaffected progress.
We present \revise, a validity-guided runtime for fine-grained recovery in structured agent workflows. When a revision arrives, \revise first intersects its delta with recorded data and control dependencies and propagates the resulting impact through the partially executed DAG to identify affected work. It then stops invalid work, preserves validity-established progress beyond the earliest conflict, and recomputes only the affected region. Incomplete provenance conservatively expands recovery, while reused results are revalidated before commit.
Analysis of real coding-agent traces show online recovery opportunities: 118 sessions retain observable work before a queued later message is delivered; across 167 overlapping assistant responses, enqueue-to-completion overlap reaches 56.55~s at p95. Across 300 challenging revision/commit executions, \revise matches a latest-version oracle with no stale outputs or effects. On unmodified LangGraph and LLMCompiler applications using Qwen3-14B, it reduces model calls by 40.6--56.0\% relative to full restart and by 31.3--43.6\% relative to suffix recomputation. Under serving pressure, it further reduces revision-to-correct-completion tokens by 13.26\% and improves SLO goodput by 3.07--5.43\%.
\end{abstract}

\section{Introduction}

Multi-turn agent workflows are becoming increasingly interactive: users can provide new instructions, corrections, or clarifications while model and tool calls from an earlier turn are still in flight~\citep{baumann2026swechat}. These inputs may change requirements, intermediate results, or control decisions on which ongoing or completed tasks depend. We call such an event a \emph{revision}. When a revision arrives during ongoing agent execution, the runtime faces a fundamental correctness--efficiency trade-off. Delaying intervention allows potentially stale in-flight work to continue, wasting computation after the revision arrives; full restart preserves latest-version correctness but discards all prior progress; and recomputing a linear suffix from the earliest conflict is more selective yet can still redo unaffected work beyond that point. The central challenge is therefore to determine fine-grained execution validity after a revision, so invalid work can be stopped while valid progress is preserved, balancing latest-version correctness with recovery efficiency.

Existing systems address different aspects of this trade-off, but do not provide a mechanism for fine-grained recovery after online revisions. Workflow and serving runtimes schedule structured model and tool execution~\citep{mei2025aios,zheng2024sglang,luo2026agentix}, but do not determine how a revision changes the validity of partially executed work. Revisable execution identifies an earliest conflict and rolls back the subsequent trace~\citep{zhai2026revisable}, but its linear recovery boundary can discard unaffected work beyond that conflict. Transactional runtimes protect externally visible effects~\citep{mohammadi2026atomix,chen2026cordon}, but do not decide which prior computation can be safely retained. What remains unresolved is a fine-grained recovery mechanism over a partially executed DAG: which work is invalidated, and which progress can still be preserved.

We present \revise, a validity-guided runtime for fine-grained recovery in structured agent workflows. When a revision arrives, \revise intersects its delta with recorded data and control dependencies and propagates the resulting impact through the executed DAG to identify affected work. It then stops invalid work, preserves proven-valid progress beyond the earliest conflict, and recomputes only the affected region. Fine-grained provenance enables selective recovery; when provenance is incomplete, \revise conservatively expands recovery toward a suffix or full restart. Reuse remains provisional: outputs and staged tool effects are revalidated against intervening revisions before commit.

Analysis of real coding-agent traces shows opportunities for online intervention: among 5,825 analyzable sessions, 118 retain observable work before a queued later message is delivered; across 167 overlapping assistant responses, enqueue-to-completion overlap reaches 56.55~s at p95. Across 300 adversarial executions, \revise matches a latest-version oracle without stale outputs or effects. On unmodified LangGraph and LLMCompiler applications using Qwen3-14B, it reduces model calls by 40.6--56.0\% relative to full restart and by 31.3--43.6\% relative to suffix recomputation. Under GPU serving pressure, it further reduces revision-to-completion tokens by 13.26\% and improves SLO goodput by 3.07--5.43\%. These results show that timely intervention avoids obsolete work, while provenance-guided selective recovery preserves additional progress when revisions are local. 

In summary, our contributions are as follows:
\begin{itemize}[leftmargin=*,topsep=2pt,itemsep=1pt]
    \item an \emph{online execution validity} abstraction for determining which completed, running, and pending work remains valid after a revision;
    \item a validity-guided, fail-closed recovery protocol that identifies affected work from revision deltas and dynamic data/control dependencies, selectively preserves valid progress, and revalidates outputs and effects before commit;
    \item an evaluation demonstrating real online revision opportunities, latest-version correctness across challenging event orderings, and lower recomputation with serving gains under contention.
\end{itemize}
\section{Revision opportunities and the limits of offline evidence}
\label{sec:motivation}

\begin{figure}[t]
  \centering
  \includegraphics[width=\linewidth]{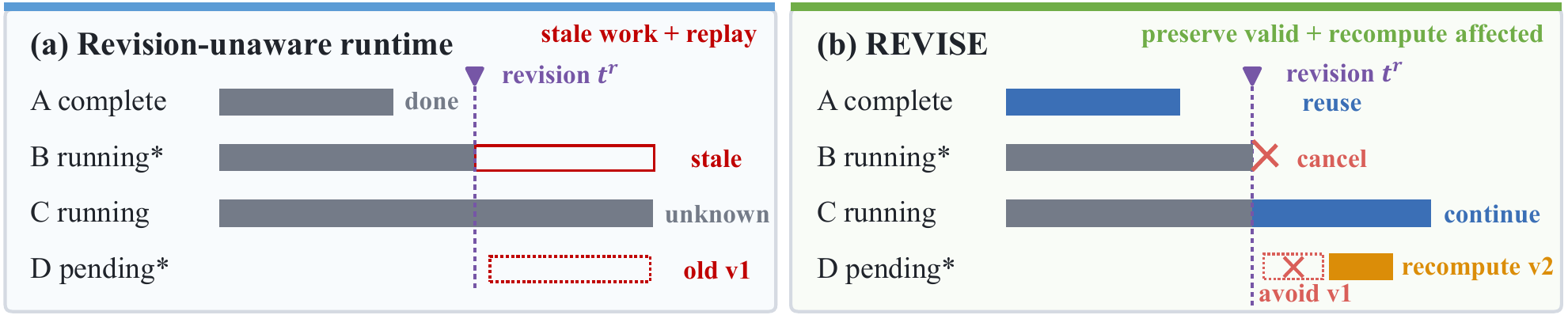}
  \caption{Desired behavior at revision time $t_r$. A revision-unaware runtime cannot safely distinguish obsolete from still-valid work. \revise aims to stop affected work, preserve valid progress, and recompute only the affected region.}
  \label{fig:concept-motivation}
\end{figure}

A revision can create two distinct recovery opportunities in a partially executed workflow. Temporal overlap allows obsolete in-flight work to be stopped early, while locality allows valid progress outside the affected region to be preserved. We therefore ask two empirical questions: \emph{Do revisions arrive early enough to intervene?} and \emph{Do real revisions exhibit locality that could permit partial recovery?} Figure~\ref{fig:concept-motivation} illustrates the desired behavior: stop obsolete work, preserve progress whose validity can be established, and recompute only the affected paths.

To answer the first question, we need both conversational semantics and an execution timeline. SWE-chat~\citep{baumann2026swechat} provides real coding-agent sessions from which we reconstruct when later user messages were queued relative to ongoing assistant, tool, and background work. Static issue-resolution benchmarks such as SWE-bench~\citep{jimenez2024swebench} cannot establish such intervention windows because they contain neither interleaved user revisions nor an execution timeline. Timing alone, however, does not imply invalidation: a later message may be an ordinary follow-up rather than a revision. We therefore combine timestamp reconstruction with manual semantic audit.

\begin{figure}[t]
  \centering
  \includegraphics[width=\linewidth]{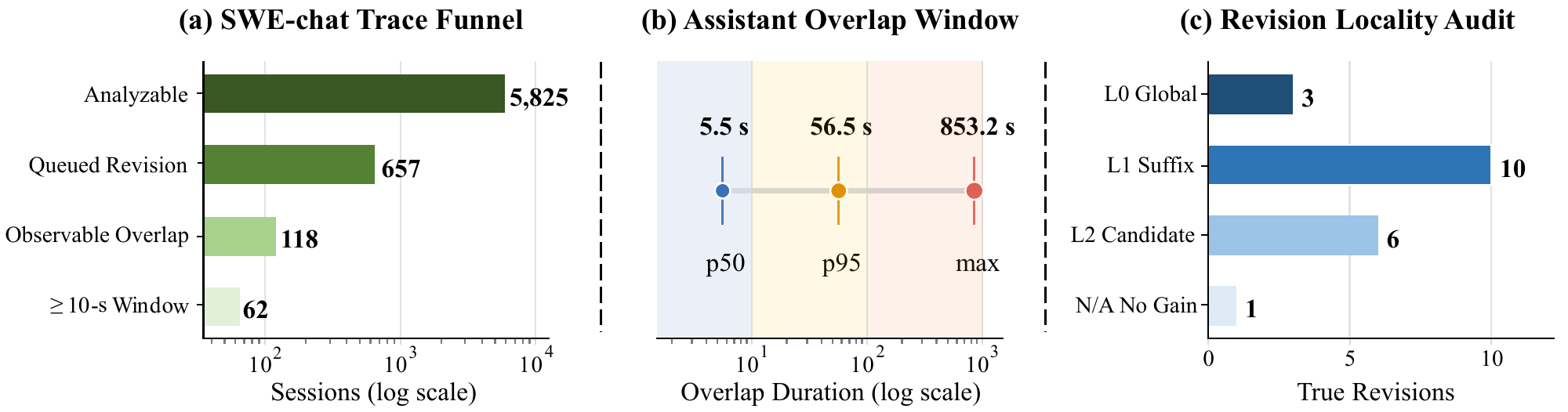}
  \caption{SWE-chat evidence for revision opportunity: (a) the trace funnel, (b) overlap duration, and (c) locality among manually audited revisions.}
  \label{fig:opportunity}
\end{figure}

The traces reveal a real but long-tailed intervention opportunity.
Among 5,825 analyzable SWE-chat sessions, 657 contain a matched queued-revision signal; 118 of these retain observable assistant, tool, or background work before the revision is delivered. These sessions contribute 174 work-bearing queued-revision events, and 62 sessions expose an intervention window of at least 10~s. Assistant responses completed within these windows overlap for 5.54~s at the median, 56.55~s at p95, and 853.16~s at the maximum. These measurements establish \emph{temporal opportunity}: revisions can arrive early enough for a runtime to stop ongoing work. They do not establish how much computation is avoidable, because the traces lack token-level progress and dependency-complete execution state.

We next examine whether these revisions appear local enough to permit partial recovery. From the 174 work-bearing events, we retain 173 with complete timestamps and deterministically select 30 for manual audit, spanning 29 sessions and 24 repositories. The sample covers all nine mutation, task-notification, or tool-overlap events together with three assistant-only duration strata; it is intended for taxonomy construction,
not prevalence estimation.

The audit identifies 20 revisions and reveals heterogeneous recovery structure: three affect a linear chain or the whole workflow (L0), ten primarily admit suffix recovery (L1), six exhibit an apparently independent branch (L2 candidate), and one has no useful intervention window. Crucially, apparent locality is not sufficient to authorize reuse. A branch that looks independent in the conversational trace may still depend on a revised artifact or control decision that the trace does not expose. SWE-chat lacks artifact versions and complete data/control provenance, so the L2 cases are only candidates for selective recovery rather than evidence that reuse is safe.

The workload study therefore establishes both sides of the motivation: revisions can overlap with ongoing agent execution, and some revisions appear local enough that coarse rollback may discard useful progress. At the same time, offline traces cannot certify which progress remains valid. This evidence gap motivates \revise's online provenance and fail-closed validity protocol.
\section{REVISE}
\label{sec:design}

\begin{figure}[t]
  \centering
  \includegraphics[width=\linewidth]{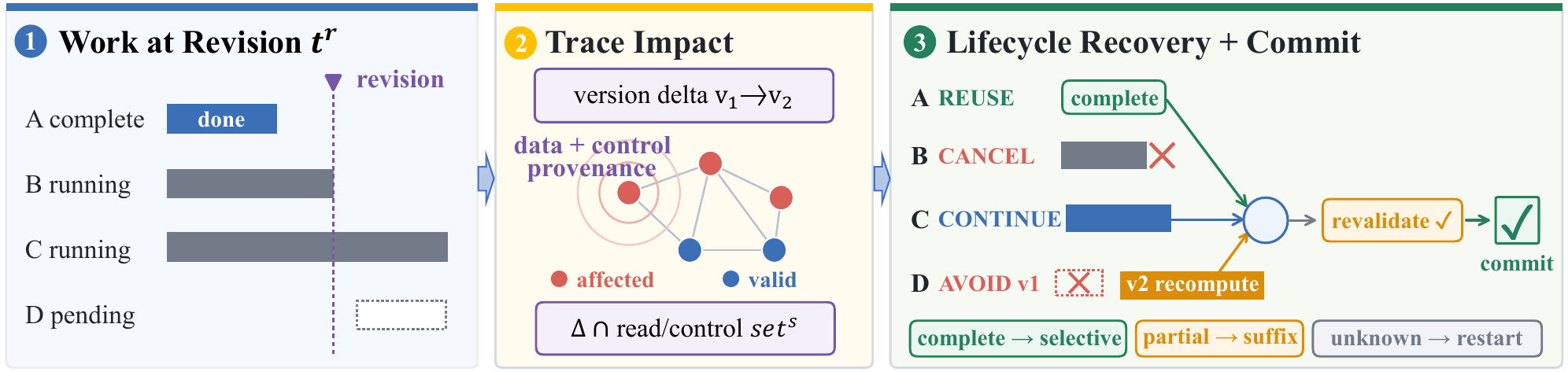}
  \caption{\revise recovery from revision to commit.  A revision arrives with completed, running, and pending work. \revise intersects its version delta with dynamic data/control provenance, maps validity to lifecycle actions---reuse, cancel, continue, avoid, or recompute---and revalidates joined outputs before commit.  Complete evidence enables selective recovery; partial or unknown evidence expands recovery to a suffix or full restart.}
  \label{fig:overview}
\end{figure}

\subsection{Validity and safe reuse}

Intuitively, an earlier attempt is reusable only when the data it read, the control decision it followed, its parent work, and any effect it would publish all remain valid.  For example, a revision to \texttt{plan.budget} invalidates an attempt that read that field, but need not invalidate an independent attempt that read neither that field nor an affected parent.

A workflow is a DAG $G=(V,E)$ whose nodes may invoke models, tools, verifiers, routers, or joins.  Each semantic artifact $a$ has a monotonically increasing version $v_a$.  A read selector $(a,p,c)$ identifies an artifact, a nested semantic path, and whether the read supplied data or controlled execution ($c\in\{\mathsf{data},\mathsf{control}\}$).  A revision event $e=(a,v,v+1,op,\Delta)$ replaces, appends, cancels, or changes control state; $\Delta$ is the set of changed paths. The runtime accepts $e$ only when $v$ is current. An attempt $x_u$ is valid if no revision since its snapshot intersects any data or control selector it consumed, every parent attempt remains current, and its effect remains authorized. The core invariant is
\begin{equation}
  \mathsf{reuse}(x_u) \Rightarrow \mathsf{valid}(x_u) \land \mathsf{effectSafe}(x_u).
\end{equation}
When either condition cannot be proved, \revise recomputes or blocks rather than guessing.

\subsection{Dynamic impact and recovery}

\revise first identifies the work that a revision can affect.  It finds recorded reads that intersect the changed paths $\Delta$.  Ordinary paths intersect under ancestor--descendant overlap.  A structural read uses a \texttt{\#members} marker to record a collection's membership, so member insertion or deletion intersects the marker only for that collection.  To collect reads without manual dependency labels, an adapter wraps structured state at node entry in a recursive read-only proxy: field access records a leaf path, whereas iteration records a membership marker.  The execution framework supplies parent edges.  A typed patch is merged into the current snapshot, and an old/new structural diff produces $\Delta$ without an application-provided affected set.

\revise indexes complete readers by artifact path, coarse readers by artifact, and unknown readers in a conservative set.  For each revision, it finds direct readers whose selectors intersect $\Delta$ and traverses their active descendants.  This produces an impact set and a recovery plan that records direct taint, affected nodes, unsafe effects, and the evidence behind every decision.

The plan assigns each node one of five lifecycle actions:
\begin{itemize}[leftmargin=*,topsep=0pt,itemsep=0pt,parsep=0pt]
  \item \emph{Cancel}: stop affected work that is currently running.
  \item \emph{Avoid}: do not start affected work that is still pending.
  \item \emph{Recompute}: execute again, on the revised snapshot, any affected completed node and any canceled or avoided node still required by the revised workflow.
  \item \emph{Continue}: let a proven-unaffected running attempt proceed; result remains provisional until commit.
  \item \emph{Reuse}: keep a proven-unaffected completed result; it too is revalidated at commit.
\end{itemize}
Thus, \emph{preserve} is not a separate action: it means continuing unaffected running work or reusing unaffected completed work.  The first three actions remove or redo stale work, while the last two retain only work supported by the available evidence.

The same evidence determines the \emph{scope} of recomputation.  Complete fine-grained provenance recomputes only the affected subgraph.  If \revise can identify only the earliest safe conflict, it recomputes a topological suffix; if it cannot identify a safe boundary, it performs a full restart.  An irreversible or incompletely covered effect requires explicit handling or blocks execution.  Selective recovery is therefore an optimization enabled by the validity semantics.

\subsection{Commit-time revalidation}

A revision may precede an attempt's final read, so revision-time decisions cannot permanently authorize reuse.  Each attempt runs on an immutable snapshot and accumulates a certificate
\begin{equation}
  C_x=(\mathit{id},v_{start},R_{data},R_{control},P,mode),
\end{equation}
where $v_{\mathrm{start}}$ is the revision-journal version at attempt start, $P$ identifies parent attempts and $mode$ is complete, coarse, or unknown. Revisions append deltas to a journal.  Revision and commit serialize on the same lock.  At commit, \revise checks every delta after $v_{start}$ against the final certificate and verifies that every parent remains the current committed attempt.  A conflicting, unknown, or obsolete attempt is rejected; tool effects remain staged until this check succeeds.  If a later revision invalidates an already committed parent, \revise invalidates its dependent outputs and any revocable or versioned effects before recomputing the current version.\footnote{\textbf{System boundary.} The recovery planner is backend independent and requires adapters only to expose workflow topology, lifecycle events, and reads. \revise decides semantic validity; physical KV allocation, including non-contiguous or cross-GPU KV reuse, remains outside its scope.}

\section{Evaluation}
\label{sec:evaluation}

\paragraph{Experimental setup.}
We evaluate \revise on LangGraph and LLMCompiler~\citep{kim2023llmcompiler},
repository-grounded replay, adversarial tests, and multi-tenant serving.
Both applications use local Qwen3-14B inference through SGLang. Serving experiments use 64 tenants and four H100 replicas at 1.4--2.6 workflows/s
(0.7--1.3$\times$ a measured 2.0-workflow/s capacity).

\paragraph{Repository-grounded replay.}
We construct controlled replays from SWE-Review-Traj~\citep{wang2026swereview}. Of 100 audited trajectories, 51 request patch changes and 33 expose structured defect paths; we select five locally reproducible workloads. We preserve the repository, patch, reviewer evidence, and tests while controlling revision timing, so the replay grounds revision content and tool work rather than natural timing.

\paragraph{Runtime and baselines.}
Our Python adapters obtain LangGraph topology/state reads and LLMCompiler
\texttt{Task} dependencies, with SGLang serving~\citep{zheng2024sglang};
they also support request aborts and staged effects. We compare
\emph{full restart}, \emph{earliest-conflict dynamic suffix}, and
\emph{\revise selective recovery}; live experiments additionally wait for turn end. Suffix receives the same provenance as \revise, isolating recovery granularity. We report correctness, calls, tokens, model-wall time, recomputation, 25-s SLO goodput, and p99 latency.

\subsection{RQ1: Does \revise preserve correctness?}

Table~\ref{tab:correctness} summarizes 15,939 executions. Every run matches its latest-version/full-restart oracle, with no stale output or committed effect. The 300-run adversarial matrix covers both revision/commit orders, late reads, consecutive revisions, unknown provenance, membership changes, and control-route invalidation. Obsolete attempts never publish, while valid siblings remain reusable; all 800 validity-certificate checks complete in under 0.04~ms. These results establish in-process validity and effect safety, not distributed consensus or crash recovery.

\begin{table}[h]
  \caption{Correctness and effect safety across complementary settings.}
  \label{tab:correctness}
  \centering
  \footnotesize
  \setlength{\tabcolsep}{2.5pt}
  \renewcommand{\arraystretch}{0.9}
  \setlength{\abovecaptionskip}{2pt}
  \setlength{\belowcaptionskip}{1pt}
  \begin{tabular}{lrrrr}
    \toprule
    Workload & Runs & Oracle-equal $\uparrow$ &
    Stale out. $\downarrow$ & Stale eff. $\downarrow$ \\
    \midrule
    Online and integration matrices & 171 & 171 & 0 & 0 \\
    LLMCompiler portability & 288 & 288 & 0 & 0 \\
    Repository-grounded replay & 180 & 180 & 0 & 0 \\
    Adversarial protocol & 300 & 300 & 0 & 0 \\
    GPU serving pressure & 15,000 & 15,000 & 0 & 0 \\
    \midrule
    Total & 15,939 & 15,939 & 0 & 0 \\
    \bottomrule
  \end{tabular}
\end{table}

\subsection{RQ2: How much recomputation is avoided?}

On LangGraph, \revise reduces model calls by 56.0\% versus full restart and 43.6\% versus dynamic suffix; model-wall time falls by 62.7\% and 50.2\%. Across 48 LLMCompiler workflows, calls fall by 40.6\% and 31.3\%, while tokens fall by 7.9\% and 5.8\%, respectively.

Because dynamic suffix uses the same provenance, these gains come from finer recovery rather than better dependency information. With unavailable fine-grained provenance, \revise falls back conservatively: all 48 cases lacking fine-grained provenance recover equivalently to full restart.

Repository replay shows the same effect on real coding artifacts. Across 18 replay configurations, \revise performs 12.94\% less recomputation than suffix (95\% CI: 7.77--17.70\%) and discards 10.38\% less completed work (95\% CI: 2.74--18.81\%). All 180 executions remain equivalent in final output, \texttt{pytest}, file/patch state, and visible effects.

\subsection{RQ3: Do work savings become serving gains?}

Work reduction need not shorten a single request's critical path. LangGraph E2E p50 improves only 5.9\% over suffix despite 50.2\% less model-wall time, while repository replay changes E2E by $-0.55$\% (95\% CI: $[-4.00,+3.84]$\%). We therefore evaluate whether saved work helps under replica contention. Across 15,000 workflows with 50\% revised requests, \revise reduces revision-to-correct-completion tokens by 13.26\% (95\% CI: 12.86--13.67\%) and model-wall time by 14.70\%
(95\% CI: 14.28--15.11\%) relative to suffix. Table~\ref{tab:pressure}
shows little goodput gain near capacity but increasing benefit under contention. Thus, work reduction is stable across load, while its serving value is pressure-dependent.

\begin{table}[!ht]
  \caption{Serving change of \revise relative to suffix.
  Brackets report 95\% confidence intervals.}
  \label{tab:pressure}
  \centering
  \footnotesize
  \setlength{\tabcolsep}{3pt}
  \begin{tabular}{llrr}
    \toprule
    Load & Regime & SLO goodput $\uparrow$ & p99 $\downarrow$ \\
    \midrule
    1.0$\times$ & Frozen capacity
      & $-0.12$\% & $-1.13$\% \\
    1.1$\times$ & Boundary pressure
      & $+3.07$\% & $-4.42$\% \\
    1.3$\times$ & Intentional overload
      & $+5.43$\% & $-13.80$\% \\
    \midrule
    Pooled & 1.0--1.3$\times$
      & $+2.79$\% $[+1.23,+4.40]$
      & $-6.45$\% $[-10.11,-3.05]$ \\
    \bottomrule
  \end{tabular}
\end{table}
\section{Related work}

\paragraph{Revision and rollback recovery.}
\emph{Revisable by Design} rolls a streaming trace back to its earliest conflict \citep{zhai2026revisable}.  Error-triggered alternatives include stepwise/Web rollback \citep{li2025garollback,zhang2026webrollback}, context--environment rewind \citep{zhuang2026agentrewind}, and DART's dependency- and effect-aware checkpoint admission \citep{yang2026dart}.  These systems roll back trajectories or checkpoints.  In contrast, \revise maps an external semantic revision to leaf-level validity decisions across node lifecycles, allowing it to preserve non-contiguous work when preservation is proved safe.

\paragraph{Foundations, substrates, and workload characterization.}
Self-adjusting computation tracks dependencies and propagates changes
\citep{anderson2021psac}; \revise extends this perspective to partially executed
agent DAGs with cancellation, staged effects, and revision/commit races.
AIOS, SGLang, and Agentix schedule agent work
\citep{mei2025aios,zheng2024sglang,luo2026agentix}; PASTE overlaps execution
\citep{paste2026}; Leyline edits KV spans \citep{ma2026leyline}; and
DeltaBox/Crab restore sandboxes \citep{deltabox2026,crab2026}.
TraceLab characterizes long-running coding-agent trajectories and human interaction gaps
\citep{zhu2026tracelab}, but does not identify whether a later interaction revises
already-consumed execution state or which completed or in-flight work remains valid afterward.
ACRFence exposes replay hazards \citep{zheng2026acrfence}, while Atomix/Cordon
protect externally visible effects \citep{mohammadi2026atomix,chen2026cordon}.
These systems provide complementary execution and recovery substrates, but do not jointly map online revisions to data/control-validity decisions over a partially executed workflow.

\section{Conclusion}

Agent workflows must remain correct when user revisions arrive during ongoing execution without unnecessarily discarding valid progress. We presented \revise, a validity-guided runtime for fine-grained recovery that identifies affected work from revision deltas and dynamic data and control dependencies, stops invalid work, selectively preserves proven-valid progress, and recomputes only the affected region. When provenance is incomplete, \revise conservatively expands recovery toward a suffix or full restart, while outputs and staged tool effects are revalidated before commit. Analysis of real SWE-chat traces establishes opportunities for online intervention, and experiments on unmodified LangGraph and LLMCompiler applications, repository-grounded replays, challenging event orderings, and Qwen3-14B serving-pressure settings show that \revise maintains latest-version correctness without stale outputs or effects, reduces unnecessary recomputation, and improves serving efficiency under contention.

{\small
\bibliographystyle{unsrtnat}
\bibliography{references}
}

\appendix
\clearpage
\section{Additional Experimental Details}
\label{sec:appendix}

\subsection{Online revision opportunity in SWE-chat}
\label{sec:audit-cohort}

The 118 SWE-chat sessions with observable work before revision delivery contain 174 work-bearing queued-revision events. We retain 173 events from 117 sessions with complete enqueue and delivery timestamps.

\paragraph{Audit-cohort construction.}
The audit includes all nine rare strict-overlap events---one mutation, two task notifications, and six tool overlaps---plus seven assistant-only events from each duration stratum: $<10$~s, 10--30~s, and $\geq30$~s. Within each stratum, a stable hash fixes the order and a greedy rule favors unused sessions and underrepresented repositories. The resulting 30-event cohort spans 29 sessions and 24 repositories. It is frozen before semantic review and is used for taxonomy construction, not prevalence estimation.

\subsection{Latest-version correctness}
\label{sec:appendix-correctness}

We evaluate latest-version correctness under challenging event orderings and repository-grounded recovery scenarios.

\paragraph{Challenging event orderings.}

\begin{table}[h]
  \caption{Challenging event orderings and required recovery behavior.}
  \centering
  \footnotesize
  \setlength{\tabcolsep}{3pt}
  \begin{tabular}{lp{0.66\linewidth}}
    \toprule
    Case & Required behavior \\
    \midrule
    Late read & A read materialized after a revision is revalidated at commit; an invalid attempt is rejected. \\
    Commit race & Commit-first output is subsequently invalidated; revision-first rejects the stale attempt at commit. \\
    Consecutive revisions & Obsolete v1/v2 attempts reject while a disjoint sibling remains valid. \\
    Unknown provenance & Missing dependency evidence fails closed. \\
    Membership change & A \texttt{\#members} read intersects key insertion or deletion. \\
    Control route & Old route certificates invalidate and the current route recomputes. \\
    \bottomrule
  \end{tabular}
\end{table}

Both possible orderings between revision and commit are exercised in the 300-run matrix. Across all runs, \revise matches the latest-version oracle without stale outputs or effects. Across 800 commit checks, certificate validation costs 0.0026~ms p50, 0.0076~ms p95, and 0.0344~ms maximum.

\paragraph{Repository-grounded correctness.}

\begin{table}[h]
  \caption{Correctness on 30 reviewer-triggered issues from 20 repositories.}
  \centering
  \footnotesize
  \setlength{\tabcolsep}{3pt}
  \begin{tabular}{lrrr}
    \toprule
    Policy & Official verdict $\uparrow$ & Latest-version equivalent $\uparrow$ & Stale effects $\downarrow$ \\
    \midrule
    Full restart & 30/30 & 30/30 & 0 \\
    Earliest-conflict suffix & 30/30 & 30/30 & 0 \\
    \textbf{\revise} & \textbf{30/30} & \textbf{30/30} & \textbf{0} \\
    \bottomrule
  \end{tabular}
\end{table}

The official verdict is joined to a reference Patch@v2, and a three-task canary reruns the official evaluator under all three policies. This experiment tests recovery correctness rather than patch quality. A separate local Qwen3-14B canary resolves 0/3 tasks, so we make no claim that \revise improves patch generation.

\subsection{Selective recovery efficiency}
\label{sec:appendix-selectivity}

We evaluate how selective recovery reduces recomputation across LangGraph, LLMCompiler, and repository-grounded replay workloads.

\paragraph{LangGraph live-request results.}

\begin{table}[h]
  \caption{Qwen3-14B work and latency in the unmodified 10-node LangGraph application.}
  \centering
  \footnotesize
  \setlength{\tabcolsep}{3pt}
  \begin{tabular}{lrrrr}
    \toprule
    Policy & Calls $\downarrow$ & Compl. tok. $\downarrow$ & Model-wall $\downarrow$ & E2E p50 $\downarrow$ \\
    \midrule
    Full restart & 10.0 & 80.0 & 1521 ms & 719 ms \\
    Earliest-conflict suffix & 7.8 & 62.4 & 1138 ms & 711 ms \\
    \textbf{\revise} & \textbf{4.4} & \textbf{35.2} & \textbf{567 ms} & \textbf{669 ms} \\
    \bottomrule
  \end{tabular}
\end{table}

The matrix delivers typed revisions during live SGLang generations and issues 15 real aborts. Policy and scenario order rotate across three repeats. Because the shared service cache is not flushed between jobs, calls and generated tokens are primary metrics; model-wall time is secondary.

\begin{table}[h]
  \caption{\revise reduction relative to earliest-conflict suffix in the live-request Qwen matrix (means over three repeats).}
  \centering
  \footnotesize
  \setlength{\tabcolsep}{3pt}
  \begin{tabular}{lrrrr}
    \toprule
    Revision & Calls & Prompt tok. & Compl. tok. & Model-wall \\
    \midrule
    Policy/control & 37.5\% & 36.3\% & 42.6\% & 43.8\% \\
    Helper goal & 28.6\% & 28.1\% & 28.6\% & 28.3\% \\
    Reviewer evidence & 16.7\% & 15.9\% & 16.7\% & 16.8\% \\
    \bottomrule
  \end{tabular}
\end{table}

\paragraph{LLMCompiler portability.}

\begin{table}[h]
  \caption{Recovery work in the unmodified LLMCompiler application \citep{kim2023llmcompiler} (48 identities per scope; means).}
  \centering
  \footnotesize
  \setlength{\tabcolsep}{3pt}
  \begin{tabular}{llrrr}
    \toprule
    Scope & Policy & Calls $\downarrow$ & Total tok. $\downarrow$ & Full fallback \\
    \midrule
    Local & Full restart & 6.00 & 2536.5 & 100\% \\
    Local & Earliest-conflict suffix & 5.19 & 2482.0 & 50\% \\
    Local & \textbf{\revise} & \textbf{3.56} & \textbf{2336.9} & \textbf{0\%} \\
    Global/untracked & All policies & 6.00 & 2552.1 & 100\% \\
    \bottomrule
  \end{tabular}
\end{table}

The portability matrix fixes eight ParallelQA plans, four revision classes, and three repeats per policy. All 48 local identities remain selective, whereas all 48 cases lacking fine-grained provenance recover equivalently to full restart. The resulting 50\% fallback rate is a controlled stress-matrix composition, not a deployment estimate. All 288 runs match the latest-version oracle with no stale outputs or effects.

A 30-pair no-revision probe produces identical outputs. Its measured $-4.2$\% wall-time difference is treated as scheduling noise rather than a speedup. Planner latency is 0.207~ms p50.

\paragraph{Repository-grounded locality.}

\begin{table}[h]
  \caption{Selective recovery in the five repository-grounded replay workloads.}
  \centering
  \footnotesize
  \setlength{\tabcolsep}{3pt}
  \begin{tabular}{llrr}
    \toprule
    Role & Locality & Affected/suffix ratio & Recompute-work change \\
    \midrule
    Multi-file local & Local positive & 0.50 & $-13.50$\% \\
    Small local & Local positive & 0.75 & $-11.18$\% \\
    Mixed boundary & Policy-equivalent & 1.00 & $-2.37$\% \\
    Global change & Negative control & 1.00 & $-0.57$\% \\
    Scope expansion & Negative control & 1.00 & $-3.94$\% \\
    \bottomrule
  \end{tabular}
\end{table}

Aggregating the first two local cases yields confidence intervals excluding zero. The three ratio-one controls have intervals crossing zero and are therefore interpreted as policy equivalence rather than performance gains. These controls distinguish the benefit of preserving proven-valid progress under local revisions from cases in which fine-grained recovery correctly collapses to the suffix boundary.

\subsection{Robustness to provenance and revision frequency}
\label{sec:appendix-robustness}

The benefit of selective recovery depends on both how often revisions occur and how precisely the runtime can establish execution validity. We vary these two factors independently.

\paragraph{Revision-frequency sensitivity.}
Across 24,000 workflows with 0/10/25/50\% revised requests, no-revision model-wall time changes by only $+0.013$\%, with a confidence interval crossing zero. At nonzero revision fractions, per-revision completion-token and model-wall reductions remain approximately 12--14\%. Aggregate benefit therefore increases with the controlled revision fraction; this sweep does not estimate production prevalence.

\paragraph{Provenance sensitivity.}
A separate 12,000-workflow provenance sweep yields recomputation-to-active-work ratios of 50.0\%, 81.7\%, and 100\% under complete, half, and unknown provenance. Half coverage falls back to full-equivalent recovery for 60\% of revised requests, while unknown provenance does so for 100\%. All runs match the latest-version oracle with no stale effects. The half-coverage mask is controlled rather than a measurement of natural provenance coverage.

In the unmodified 10-node LangGraph application, complete provenance captures all six expected leaf reads and recovers all five expected recomputation sets without full-restart fallback. As provenance becomes less precise, recovery expands conservatively rather than reusing work whose validity cannot be established.

\subsection{Serving under GPU pressure}
\label{sec:appendix-serving}

The serving matrix runs Qwen3-14B with SGLang~0.5.13 on four single-H100 replicas and 64 tenant namespaces under open-loop arrivals. A separate suffix-only scan fixes measured capacity at 2.0 workflows/s before policy comparison. We evaluate 1.4, 1.8, 2.0, 2.2, and 2.6 workflows/s, corresponding to 0.7--1.3$\times$ measured capacity; the final rate is intentional overload.

Each policy runs 500 workflows per configuration (180 warmup, 280 steady, 40 cooldown) in three paired repeats, yielding 15,000 records. Job identity, arrival, tenant, revision, and replica assignment are paired, while policy order rotates across repeats. All runs use four replicas, satisfy telemetry and arrival-generator checks, and commit no stale effects. Work deltas and configuration-level serving deltas are bootstrapped separately.

\subsection{Implementation and reproducibility}
\label{sec:appendix-reproducibility}

\paragraph{Framework integration.}
The unmodified 10-node LangGraph application uses a 25-line shared state/node-factory wrapper that captures all six expected leaf reads, with no application-node edits or selector annotations. For LLMCompiler, a shared 343-logical-line adapter wraps native task arguments and dependencies with no application edits or selector annotations, capturing all 1,584 expected explicit selector instances across the portability matrix.

\paragraph{Environment and artifacts.}
Experiments use Python~3.12, LangGraph~1.2.9, SGLang~0.5.13, Qwen3-14B in bfloat16, and H100 80GB GPUs. Generation uses temperature zero and fixed experiment-order seeds. Formal runners emit per-job JSON records containing configuration, selected and executed nodes, token/model-wall metrics, effect state, and oracle checks; separate analyzers enforce frozen gates. Repository replay additionally records rootfs identity, test collection, and exit status. The anonymized artifact package includes the runners, analyzers, preregistrations, and environment instructions.

% \clearpage
% \input{checklist.tex}

\end{document}